\pdfoutput=1

\documentclass[11pt]{article}

\usepackage{graphicx}
\usepackage{booktabs}
\usepackage{subcaption}
\usepackage{makecell}

\usepackage{acl}

\usepackage{amsmath}
\usepackage{amssymb}
\usepackage{mathtools}
\usepackage{amsthm}

\usepackage{times}
\usepackage{latexsym}

\usepackage[T1]{fontenc}
\usepackage[utf8]{inputenc}

\usepackage{microtype}

\usepackage{inconsolata}

\title{Proxy-RLHF: Decoupling Generation and Alignment in Large Language Model with Proxy }

\author{Yu Zhu$^{1,2}$$\thanks{~~Equal contribution}$, Chuxiong Sun$^{4*}$, Wenfei Yang$^{1,2}$, Wenqiang Wei$^3$, Bo Tang$^3$$\thanks{~~Corresponding authors}$, Tianzhu Zhang$^{1,2\thefootnote}$\\
\bf Zhiyu Li$^3$, Shifeng Zhang$^5$, Feiyu Xiong$^3$, Jie Hu$^4$, Mingchuan Yang$^4$\\
$^1$University of Science and Technology of China, Hefei, China \\
$^2$Deep Space Exploration Laboratory \\
$^3$Institute for Advanced Algorithms Research, Shanghai, China \\
$^4$Research Institute of China Telecom \\
$^5$Sangfor Technologies Inc.
 }

\begin{document}
\maketitle
\begin{abstract}
Reinforcement Learning from Human Feedback (RLHF) is the prevailing approach to ensure Large Language Models (LLMs) align with human values. However, existing RLHF methods require a high computational cost, one main reason being that RLHF assigns both the generation and alignment tasks to the LLM simultaneously. In this paper, we introduce Proxy-RLHF, which decouples the generation and alignment processes of LLMs, achieving alignment with human values at a much lower computational cost. We start with a novel Markov Decision Process (MDP) designed for the alignment process and employ Reinforcement Learning (RL) to train a streamlined proxy model that oversees the token generation of the LLM, without altering the LLM itself. Experiments show that our method achieves a comparable level of alignment with only 1\% of the training parameters of other methods.
\end{abstract}

\section{Introduction}

Large language models (LLMs) have demonstrated formidable capabilities in various tasks including summarization~\cite{stiennon2020learning,koh2022empirical}, instruction following~\cite{chung2022scaling,ouyang2022training}, robotics~\cite{huang2023voxposer,liu2023llm+}, and more\cite{roziere2023code}. attracting widespread attention in academia and industry. 

\begin{figure}[ht]
\vskip 0.2in
\begin{center}
\centerline{\includegraphics[width=\columnwidth]{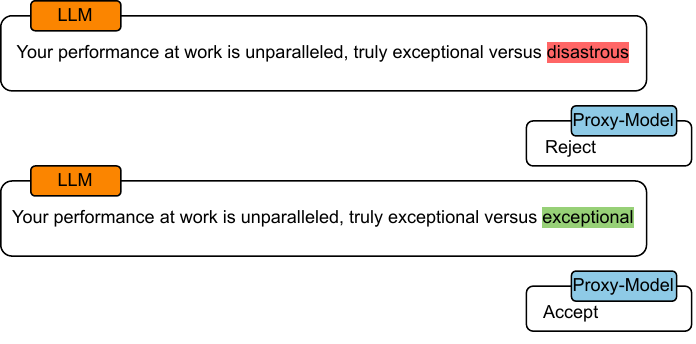}}
\caption{Demonstration of how the proxy model works. The proxy model is responsible for supervising the generation of the LLM, deciding whether to accept the latest token generated by the LLM. By accepting tokens that align with human values and rejecting those that do not, it ensures that the final generation results are aligned with human values.}
\label{mdp}
\end{center}
\vskip -0.2in
\end{figure}

Several methods~\cite{ouyang2022training,bai2022constitutional,dong2023raft,yuan2023rrhf,rafailov2023direct,dai2023safe} have been proposed to ensure the outputs of LLMs to align with human values, among which Reinforcement Learning from Human Feedback (RLHF)~\cite{christiano2017deep,ziegler2019fine,casper2023open} is the mainstream. 
RLHF models the generation process of LLMs as a Markov Decision Process (MDP) and treats the language model as a policy model, directly optimizing its parameters. 

In the RLHF method, LLMs are responsible for both generation and alignment, making the alignment process inevitably computation-intensive.  
The RLHF method employs on-policy reinforcement learning algorithms, typically PPO~\cite{schulman2017proximal}, which requires two trainable language models of the same size. Furthermore, an extra constraint on the KL divergence with the reference model is imposed.
Overall, the RLHF method requires the simultaneous use of four models—policy, reward, value, and reference models—each with billions of parameters.

To address these issues, we propose Proxy-RLHF, which aligns language models with human values with minimal computational cost. 
Different from previous methods, our core idea is to decouple the generation and alignment processes of LLMs. Specifically, we have restructured the Markov Decision Process in RLHF. In this framework, LLMs are solely responsible for generating tokens without having to consider alignment with human values. A new proxy model evaluates the quality of the generated tokens, accepting those that align with human values and rejecting those that do not, thereby achieving alignment. 

However, training a proxy model from scratch is challenging. 
Unlike the RLHF approach, where the policy model benefits from initialization with LLMs, the proxy model lacks initial understanding about natural language. 
Therefore, we propose the Stable Knowledge-Aware Module (SKAM), which can (1) stabilize training and avoid unnecessary repeated exploration through the redesign of LLMs sampling, and (2) ensure that the final generated responses fall within the knowledge and skill scope of the LLMs by limiting the rejection actions of the proxy model, potentially endowing the proxy model with certain linguistic capabilities.
Additionally, we utilize the hidden states generated by the LLMs during its generation process as input features for the proxy model, further reducing the number of parameters and computational cost.

We encapsulate the generation of LLMs into a reinforcement learning environment and conduct extensive experiments on it. The experiments validated that our method is both parameter-efficient and data-efficient, achieving a comparable level of alignment with less than 1\% of the training parameters used by other methods.

\section{Proxy-RLHF}

\subsection{Markov decision process}
We conceptualize the alignment process of large language models as a Markov Decision Process (MDP), represented by a tuple $(\mathcal{S,A,R},\mathcal{P},\pi)$, where $s \in \mathcal{S}$ denotes the state, $a \in \mathcal{A}$ represents the action, $\mathcal{P}:\mathcal{S} \times \mathcal{A} \times \mathcal{S} \to [0,1]$ signifies the state transition probability, $\mathcal{R}$ denotes the reward, and a policy $\pi:\mathcal{S} \to \mathcal{A}$ represents a mapping from state space to action space. 

Specifically, in Proxy-RLHF, let $\pi_\theta$ denotes the policy of proxy model, its input state $s$ is a sequence of tokens consisting of a prompt and the responses generated up to that point. Its action space $\mathcal{A}$ contains two actions: $a=0$ for accepting the token, and $a=1$ for rejecting. If the newly generated token is accepted, the language model generates a new candidate token based on the prefix, otherwise, it resamples a new token based on the prefix without the rejected token. That is, the state transition $\mathcal{P}$ is determined by the generation process of the LLMs, emphasizing the importance of designing a sampling method for the LLMs that facilitates the learning of the proxy model.

\subsection{Stable Knowledge-Aware Module}

The Stable Knowledge-Aware Module consists of two parts: the redesign of the sampling method and the restriction of the action space of the proxy model. The redesign of the sampling method reduces the randomness of state transitions and unnecessary exploration in the environment, stabilizing the model's training. The restriction of the proxy model's action space, by limiting the number of rejections, ensures that the final generated answers fall within the knowledge and skill range of the LLMs, guaranteeing the usefulness of the answers.

\paragraph{Redesign of Sampling}We have the LLM generate tokens in descending order of probability and remove any token that has been rejected by the proxy model at the same position from the pool of candidate tokens. This resembles greedy sampling, but the difference lies in that the same position may undergo multiple regenerations in Proxy-RLHF if previous tokens are rejected by the proxy model, meaning the final accepted token is the one with the highest probability among the remaining candidate tokens, not necessarily the highest probability of all tokens. Specifically, a new candidate token $t_i$ in step $i$ is generated by
\[
t_i = \underset{t \in \mathcal{T}\setminus \mathcal{T}^r}{\operatorname*{\text{argmax }}} p_\phi\{t|x,y_{<i}\}
\]
where $p_\phi$ represents the logits generated by the language model based on the prompt $x$ and previously generated responses $y_{<i}$. $\mathcal{T}$ is the set of the entire vocabulary. $\mathcal{T}^r$ is the set maintaining tokens rejected by the proxy model at position $i$ and will be reset to $\emptyset$ if stepping into $i+1$.

Given state $s = (x , y_{<i+1})$, we have next state
\[
s' = 
\begin{cases} 
  (x , y_{<i+1}, t_{i+1}) & \text{if } a = 0, \\
  (x , y_{<i}, t_{i}) & \text{if } a = 1.
\end{cases}
\]
Note that selecting the token greedily does not hurt the output diversity: we can always reach the desired token by consistently rejecting others. Additionally, removing rejected tokens from the candidate tokens set can prevent the training process from falling into unnecessary loops, e.g. generating and rejecting the same token again.

\paragraph{Action Space Restriction}In proxy RLHF, the probability of token $\widetilde{t}$ being accepted at position $i$ is:
\[
p\{y_i = \widetilde{t} | x,y_{<i}\}=\prod_{t_i \in \mathcal{\widetilde{T}}}\pi_\theta(a = 1|x,y_{<i},t_i).
\]
where $\mathcal{\widetilde{T}}=\{t_i|p_\phi\{t_i|x,y_{<i}\}>p_\phi\{\widetilde{t}|x,y_{<i}\},t_i \in \mathcal{T}\}$ is the set of all tokens whose probabilities, as provided by the language model's logits, are greater than $\widetilde{t}$.
The chosen probability of $\widetilde{t}$ shifts from the origin probability of the language model to the product of the action probability of the proxy model. This implies that tokens with a low probability in the language model might be chosen by the proxy model with a higher probability, creating a discrepancy.
Therefore, relying solely on existing methods of limiting LLMs generation sampling (e.g., topk, topp sampling) is no longer effective. 

To address this issue, we restrict the action space of the proxy model. Specifically, We preset a hyper-parameter $p_t$. If the average probability of the remaining tokens is less than $p_t$, we mask the rejection action, thus forcing the action of the proxy model to be acceptance. This ensures that irrational tokens are not sampled. The constrained action space of the proxy can be represented as:
\[
\mathcal{A}=
\begin{cases} 
  \{0\} & \text{if } \sum_{t \in \mathcal{T'}}p_\phi \{t|x,y_{<i}\}\leq p_t *|\mathcal{T'}|, \\
  \{0,1\} & \text{else }.
\end{cases}
\]
Where $\mathcal{T'} = \mathcal{T}\setminus \mathcal{T}^r$. In actual deployment, we use topp sampling methods with temperature to eliminate most irrational tokens beforehand, further reducing unnecessary exploration by the proxy model and stabilizing and speeding up the training process.

\section{Experiment}

We designed experiments to answer the following questions: 1. Can our method perform comparably to RLHF or DPO with far fewer training parameters? 2. How does the Stable Knowledge-Aware Module affect the performance of our method? 3. As a method trained from scratch, how data-efficient is our approach?

\begin{figure}[htb]
    \centering
    \begin{subfigure}[b]{0.22\textwidth}
        \centering
        \includegraphics[width=\textwidth]{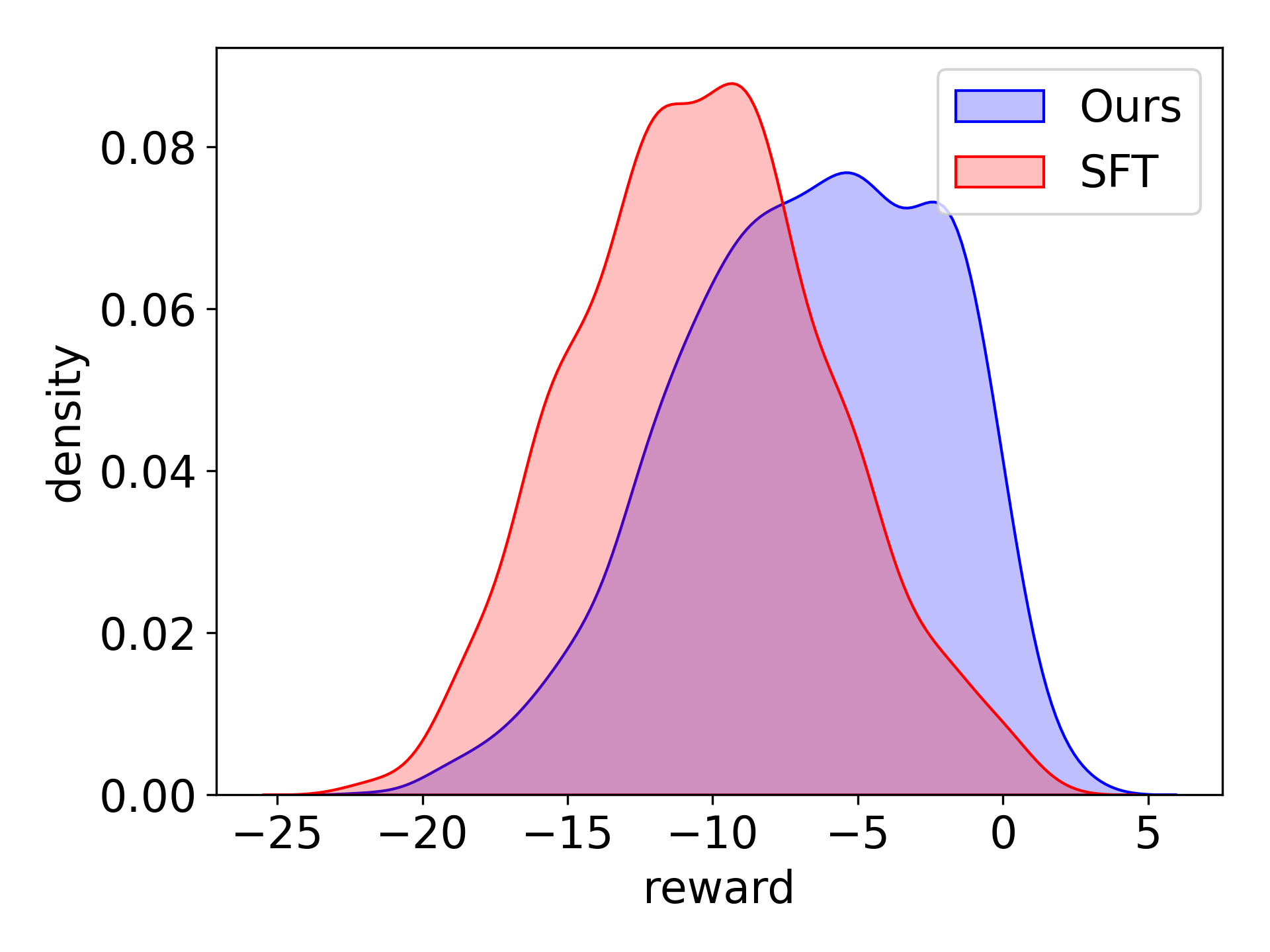}
        \caption{}
        \label{w1}
    \end{subfigure}
    \hfill
    \begin{subfigure}[b]{0.22\textwidth}
        \centering
        \includegraphics[width=\textwidth]{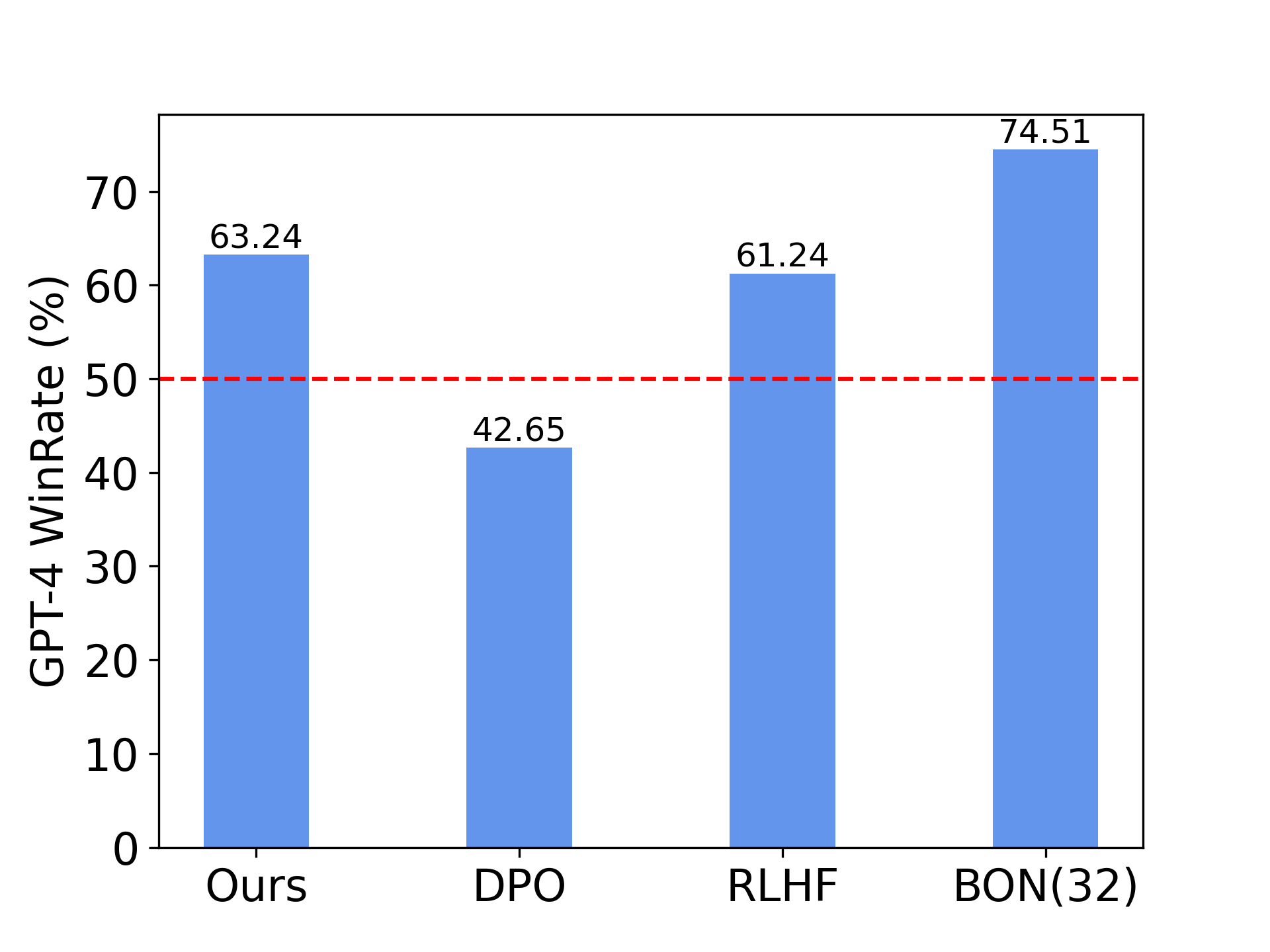}
        \caption{}
        \label{w2}
    \end{subfigure}
    \caption{(a) The reward distribution on the test set for SFT and Ours, where scores are obtained from the reward model. (b) The win rate of Ours, DPO, RLHF, and BON against the SFT model, where the win rate is determined by pair-wise comparison from GPT-4. We use greedy sampling for all methods above and set n=32 in BON.}
    \label{w}
\end{figure}

\begin{table}[t]
\centering
\caption{The comparison of the number of parameters between our method, DPO, and RLHF when fine-tuning the Alpaca-7B model}
\resizebox{\columnwidth}{!}{
\begin{tabular}{l|ccc}
\toprule
Method              &  \thead{Trainable \\ parameters}& \thead{GPU memory \\ required for training}  & \thead{Fine-tuning \\ LLM} \\
\midrule
RLHF    & 13.35B & 198.87Gb       &    \checkmark     \\
DPO    & 6.74B        & 100.41Gb   &    \checkmark  \\
\textbf{Proxy-RLHF(ours)} & \textbf{0.03B}        & \textbf{0.5Gb}       &   \\
\bottomrule
\end{tabular}
}
\label{table:canshu}
\end{table}

\paragraph{Prompt Dataset and Model}We use the hh dataset and filter out the safety-related prompts to focus more on helpfulness and reduce bias. Furthermore, similar to previous work\cite{dong2023raft}, to reduce the use of GPU memory, we do not use prompts exceeding 256 tokens in length. The final dataset comprises a training set of 36k prompts and a test set of 1899 prompts.

Consistent with previous work\cite{dai2023safe}, the experiments were conducted on the alpaca-7b model, which was obtained by applying supervised fine-tuning (SFT) to the llama-7b model using prompt data from GPT-3.5.

\paragraph{Evaluation Metrics}Two methods were used to evaluate the model's output: the score of the reward model and the win rate of GPT-4\cite{achiam2023gpt}. We use the beaver reward model\footnote{https://huggingface.co/PKU-Alignment/beaver-7b-v1.0-reward}, which was trained on the same dataset and achieved an accuracy of 62.03\% on the test set. We use GPT-4 (gpt-4-1106-preview) for pairwise comparison of the outputs to obtain the final win rate.

\begin{figure}[tb]
    \centering
    \begin{subfigure}[b]{0.22\textwidth}
        \centering
        \includegraphics[width=\textwidth]{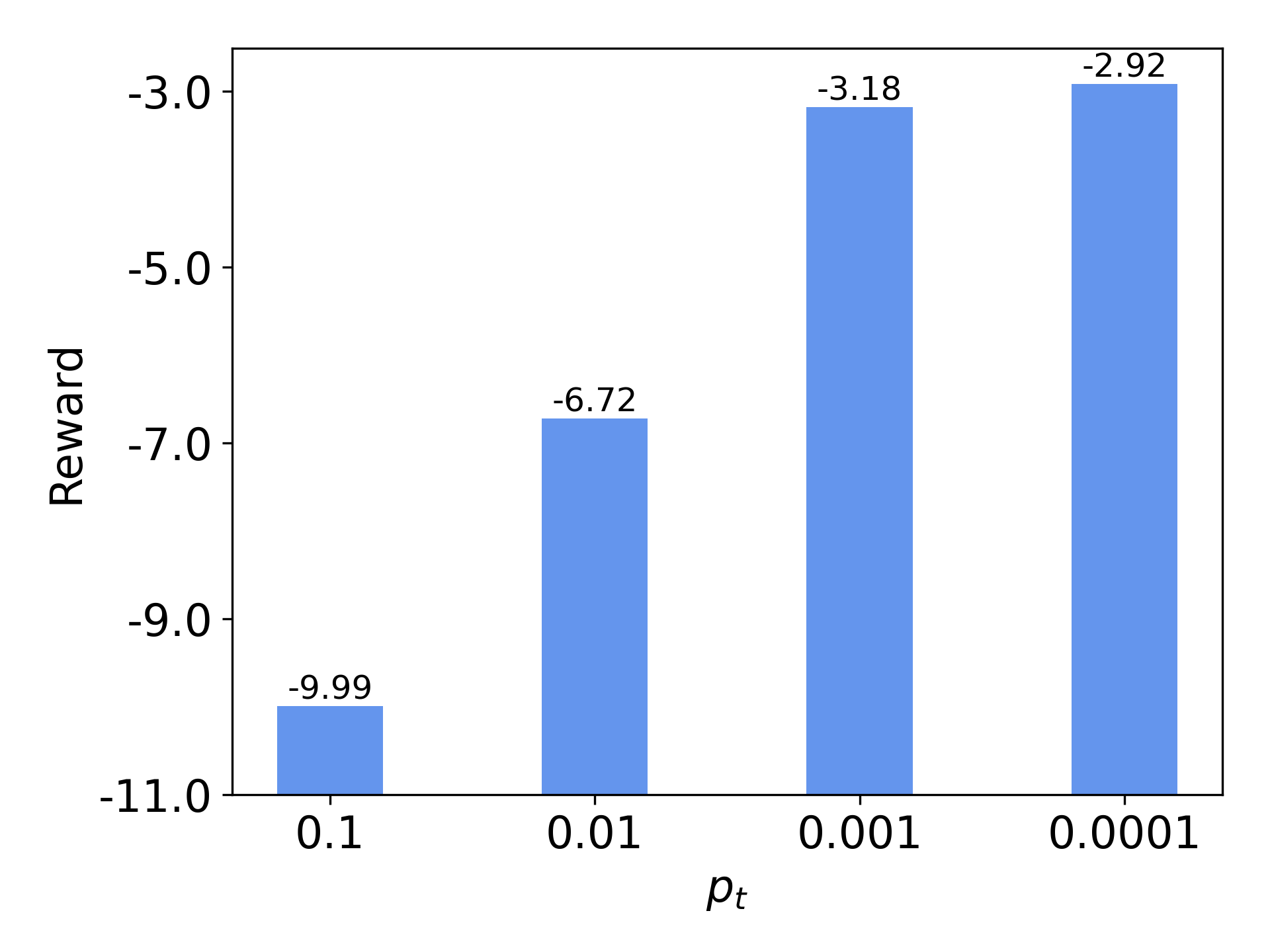}
        \caption{}
    \end{subfigure}
    \hfill
    \begin{subfigure}[b]{0.22\textwidth}
        \centering
        \includegraphics[width=\textwidth]{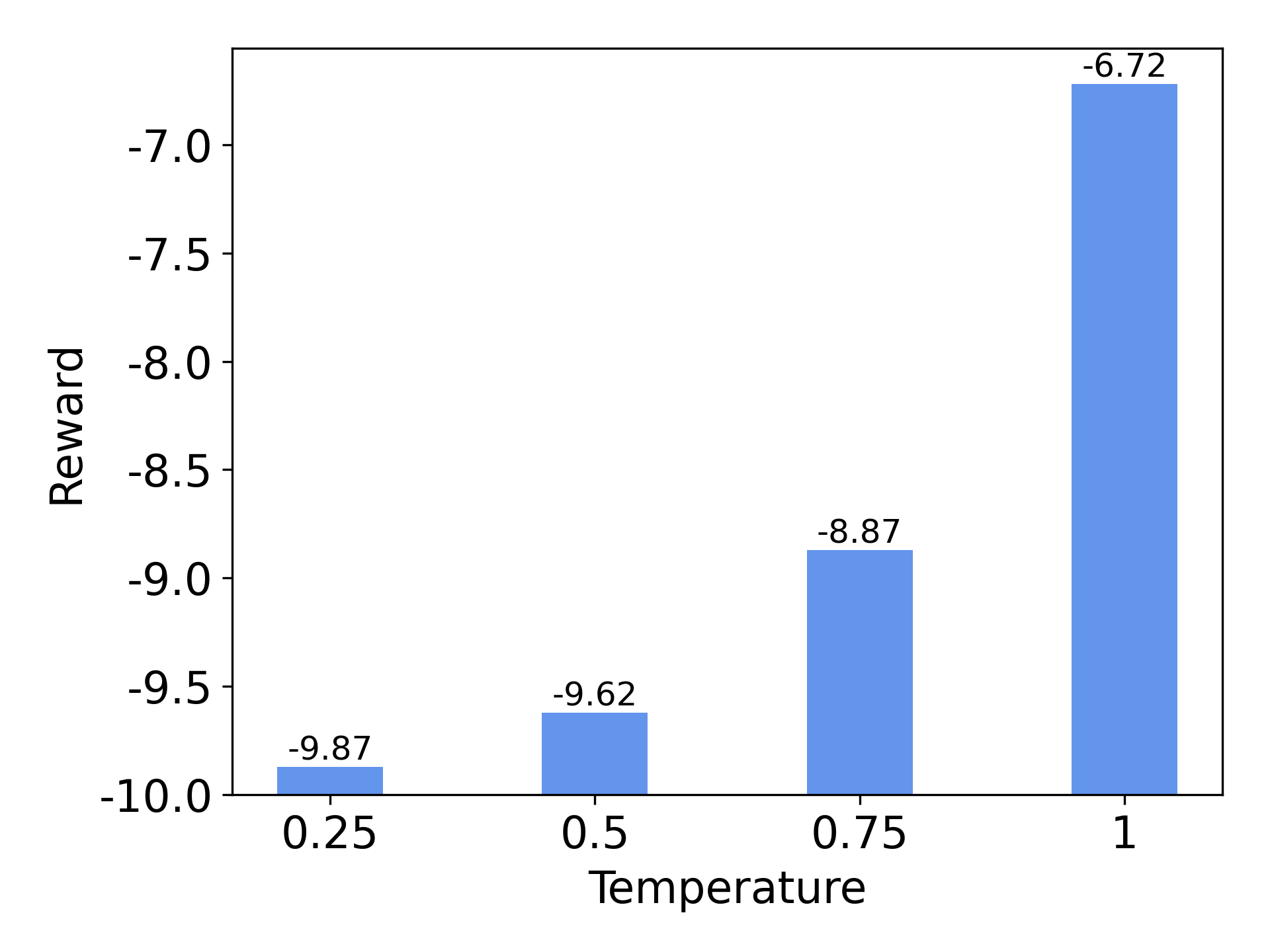}
        \caption{}
    \end{subfigure}
    \caption{(a) The average score given by the reward model on the test set, for models corresponding to different $p_t$, after completing one round on the training set. (b) The average score corresponding to different temperatures, after completing one round on the training set.}
    \label{fig:hyperparameters}
\end{figure}

\paragraph{Effectiveness Experiment}In this section, we demonstrate the effectiveness of our method, addressing question 1.

Figure \ref{w1} shows the reward distribution of SFT baseline and our method on the test set of filtered HH dataset. A significant right-shift of the distribution can be observed, which indicates that our method can effectively improve the outputs' reward of the SFT model. On the other hand, our method has higher win rate (63.24\%) of GPT-4 versus SFT baseline than DPO (42.65\%) and RLHF (61.24\%), as shown in Figure \ref{w2}. It demonstrates that our method can achieve comparable performance to RLHF and DPO with less than 1\% trainable parameters showing in Table \ref{table:canshu}. The low win-rate of DPO with greedy sampling is also validated in its paper, while our method can still achieve win-rate higher than 50\%. BON (32) achieves the best results, however, with many times of generation in inference. Like RLHF and DPO, our method only requires once generation in inference.

\begin{table}[t]
\centering
\caption{The average score on the test set for models with different temperatures on the first 0.5k, 1k, 1.5k, 2k and 36k (full) train data.}
\resizebox{\columnwidth}{!}{
\begin{tabular}{l|ccccc}
\toprule
$p_t$              & 0.5k & 1k & 1.5k & 2k & 36k \\
\midrule
0.1    & -10.14 & -10.15       & -10.12  & -9.70    &    -9.99     \\
0.01    & -7.09        & -6.53        & -6.96  &-7.05  &     -6.72   \\
0.001 & -3.93        & -4.72        & -3.70  &-4.05    &  -3.18  \\
0.0001  & -3.46       & -4.51        & -3.28  &-4.45  &     -2.92   \\
\bottomrule
\end{tabular}
}
\label{table:t1}
\end{table}

\paragraph{Hyper-parameters Experiment}In this section, we answer question 2. We demonstrate how two key hyper-parameters of the Stable Knowledge-Aware Module influence the final performance of the model. The two hyper-parameters are $p_t$ and temperature. A higher $p_t$ indicates a greater likelihood that the model's actions are restricted to only accepting the current generation result. A higher temperature means the logits output by the LLMs are processed more smoothly, leading to a more uniform probability distribution of tokens in the vocabulary and a smaller chance that the model's actions are restricted to only accepting the current generation result. In summary, the smaller the $p_t$ and the higher the temperature, the less likely it is that the model's action space is restricted, allowing for more choices, and vice versa.

The results in Figure \ref{fig:hyperparameters} show that as $p_t$ decreases and temperature increases, the reward also increases. This suggests that when the model has more choices, our proxy model can adapt to different choices and produce more high-quality outputs.
Moreover, too small values of $p_t$ (0.001 and 0.0001) may result in irregular outputs (see in Appendix). Thus, the Stable Knowledge-Aware Module is necessary for avoiding irregular outputs but it can also be tuned for proxy model to search in a larger response space and find better responses.

\paragraph{Efficient Experiment}In this section, we address question 3: We present the average scores of the model on the test set after being trained with 0.5k, 1k, 1.5k, 2k training data points, as well as the difference in average scores after one round of training with all the data in the training set.

Table \ref{table:t1} shows that the scores in early training steps are close to the final scores after one round of training. Especially for $p_t=0.01$, we can achieve the final performance in early training steps no more than 2000. This suggests that our method can quickly converge and is data efficient.

\section{Related Work}

Several approaches have been proposed to reduce the complexity and instability of RLHF\cite{ramamurthy2022reinforcement}. \cite{bai2022constitutional,lee2023rlaif} introduced RLAIF, which reduces the annotation cost of preference data. RRHF\cite{yuan2023rrhf} and RAFT\cite{dong2023raft} rank responses and use the highest-scoring answers for supervised fine-tuning. Direct Preference Optimization (DPO)\cite{rafailov2023direct} directly optimizes the language model with preference loss without the need for additional training of a reward model. Methods above consider the LLM itself as a policy model to be optimized, taking on both generation and alignment tasks, making the computationally expensive step of fine-tuning the LLM unavoidable.

\section{Conclusion}
In this paper, we introduce the proxy-model, which decouples the generation and alignment processes within LLMs, using an additional lightweight proxy model to guide the generation of LLMs, achieving an alignment of output answers with human values. Furthermore, we propose SKAM to stabilize the training of the proxy model and ensure the effectiveness of the answers. Experiments show that our method achieves a level of alignment comparable to RLHF with less than 1\% of the training parameters.

\section*{Limitations}

This study, while pioneering in its approach to decouple generation and alignment processes in LLMs, is subject to several limitations. First, the effectiveness of the Proxy-RLHF model relies heavily on the quality and comprehensiveness of human feedback, which may not always be consistent or universally applicable across different domains or cultures. Secondly, the proposed method has been primarily validated in controlled experimental settings, and its robustness in real-world applications remains to be extensively tested. Lastly, the scalability of this approach to even larger models or more complex tasks is not fully explored, leaving open questions about its long-term applicability and adaptability.

\section*{Ethics Statement}
In developing Proxy-RLHF, we recognize the ethical implications associated with the deployment of large language models (LLMs). Our method aims to align LLMs more closely with human values through efficient and targeted feedback, addressing concerns related to bias, misinformation, and the potential for harmful outputs. However, we acknowledge that the technology could be misused if the alignment process is biased or if the proxy model is manipulated to endorse unethical values. We committed to transparency in our methodology and results to foster an open dialogue about these challenges. We also emphasize the importance of diverse and inclusive feedback to mitigate biases. Moving forward, we encourage continued ethical scrutiny and multidisciplinary collaboration to ensure that advancements in LLMs contribute positively to society.

\bibliography{main}

\begin{thebibliography}{19}
\expandafter\ifx\csname natexlab\endcsname\relax\def\natexlab#1{#1}\fi

\bibitem[{Achiam et~al.(2023)Achiam, Adler, Agarwal, Ahmad, Akkaya, Aleman, Almeida, Altenschmidt, Altman, Anadkat et~al.}]{achiam2023gpt}
Josh Achiam, Steven Adler, Sandhini Agarwal, Lama Ahmad, Ilge Akkaya, Florencia~Leoni Aleman, Diogo Almeida, Janko Altenschmidt, Sam Altman, Shyamal Anadkat, et~al. 2023.
\newblock Gpt-4 technical report.
\newblock \emph{arXiv preprint arXiv:2303.08774}.

\bibitem[{Bai et~al.(2022)Bai, Kadavath, Kundu, Askell, Kernion, Jones, Chen, Goldie, Mirhoseini, McKinnon et~al.}]{bai2022constitutional}
Yuntao Bai, Saurav Kadavath, Sandipan Kundu, Amanda Askell, Jackson Kernion, Andy Jones, Anna Chen, Anna Goldie, Azalia Mirhoseini, Cameron McKinnon, et~al. 2022.
\newblock Constitutional ai: Harmlessness from ai feedback.
\newblock \emph{arXiv preprint arXiv:2212.08073}.

\bibitem[{Casper et~al.(2023)Casper, Davies, Shi, Gilbert, Scheurer, Rando, Freedman, Korbak, Lindner, Freire et~al.}]{casper2023open}
Stephen Casper, Xander Davies, Claudia Shi, Thomas~Krendl Gilbert, J{\'e}r{\'e}my Scheurer, Javier Rando, Rachel Freedman, Tomasz Korbak, David Lindner, Pedro Freire, et~al. 2023.
\newblock Open problems and fundamental limitations of reinforcement learning from human feedback.
\newblock \emph{arXiv preprint arXiv:2307.15217}.

\bibitem[{Christiano et~al.(2017)Christiano, Leike, Brown, Martic, Legg, and Amodei}]{christiano2017deep}
Paul~F Christiano, Jan Leike, Tom Brown, Miljan Martic, Shane Legg, and Dario Amodei. 2017.
\newblock Deep reinforcement learning from human preferences.
\newblock \emph{Advances in neural information processing systems}, 30.

\bibitem[{Chung et~al.(2022)Chung, Hou, Longpre, Zoph, Tay, Fedus, Li, Wang, Dehghani, Brahma et~al.}]{chung2022scaling}
Hyung~Won Chung, Le~Hou, Shayne Longpre, Barret Zoph, Yi~Tay, William Fedus, Yunxuan Li, Xuezhi Wang, Mostafa Dehghani, Siddhartha Brahma, et~al. 2022.
\newblock Scaling instruction-finetuned language models.
\newblock \emph{arXiv preprint arXiv:2210.11416}.

\bibitem[{Dai et~al.(2023)Dai, Pan, Sun, Ji, Xu, Liu, Wang, and Yang}]{dai2023safe}
Josef Dai, Xuehai Pan, Ruiyang Sun, Jiaming Ji, Xinbo Xu, Mickel Liu, Yizhou Wang, and Yaodong Yang. 2023.
\newblock Safe rlhf: Safe reinforcement learning from human feedback.
\newblock \emph{arXiv preprint arXiv:2310.12773}.

\bibitem[{Dong et~al.(2023)Dong, Xiong, Goyal, Pan, Diao, Zhang, Shum, and Zhang}]{dong2023raft}
Hanze Dong, Wei Xiong, Deepanshu Goyal, Rui Pan, Shizhe Diao, Jipeng Zhang, Kashun Shum, and Tong Zhang. 2023.
\newblock Raft: Reward ranked finetuning for generative foundation model alignment.
\newblock \emph{arXiv preprint arXiv:2304.06767}.

\bibitem[{Huang et~al.(2023)Huang, Wang, Zhang, Li, Wu, and Fei-Fei}]{huang2023voxposer}
Wenlong Huang, Chen Wang, Ruohan Zhang, Yunzhu Li, Jiajun Wu, and Li~Fei-Fei. 2023.
\newblock Voxposer: Composable 3d value maps for robotic manipulation with language models.
\newblock \emph{arXiv preprint arXiv:2307.05973}.

\bibitem[{Koh et~al.(2022)Koh, Ju, Liu, and Pan}]{koh2022empirical}
Huan~Yee Koh, Jiaxin Ju, Ming Liu, and Shirui Pan. 2022.
\newblock An empirical survey on long document summarization: Datasets, models, and metrics.
\newblock \emph{ACM computing surveys}, 55(8):1--35.

\bibitem[{Lee et~al.(2023)Lee, Phatale, Mansoor, Lu, Mesnard, Bishop, Carbune, and Rastogi}]{lee2023rlaif}
Harrison Lee, Samrat Phatale, Hassan Mansoor, Kellie Lu, Thomas Mesnard, Colton Bishop, Victor Carbune, and Abhinav Rastogi. 2023.
\newblock Rlaif: Scaling reinforcement learning from human feedback with ai feedback.
\newblock \emph{arXiv preprint arXiv:2309.00267}.

\bibitem[{Liu et~al.(2023)Liu, Jiang, Zhang, Liu, Zhang, Biswas, and Stone}]{liu2023llm+}
Bo~Liu, Yuqian Jiang, Xiaohan Zhang, Qiang Liu, Shiqi Zhang, Joydeep Biswas, and Peter Stone. 2023.
\newblock Llm+ p: Empowering large language models with optimal planning proficiency.
\newblock \emph{arXiv preprint arXiv:2304.11477}.

\bibitem[{Ouyang et~al.(2022)Ouyang, Wu, Jiang, Almeida, Wainwright, Mishkin, Zhang, Agarwal, Slama, Ray et~al.}]{ouyang2022training}
Long Ouyang, Jeffrey Wu, Xu~Jiang, Diogo Almeida, Carroll Wainwright, Pamela Mishkin, Chong Zhang, Sandhini Agarwal, Katarina Slama, Alex Ray, et~al. 2022.
\newblock Training language models to follow instructions with human feedback.
\newblock \emph{Advances in Neural Information Processing Systems}, 35:27730--27744.

\bibitem[{Rafailov et~al.(2023)Rafailov, Sharma, Mitchell, Ermon, Manning, and Finn}]{rafailov2023direct}
Rafael Rafailov, Archit Sharma, Eric Mitchell, Stefano Ermon, Christopher~D Manning, and Chelsea Finn. 2023.
\newblock Direct preference optimization: Your language model is secretly a reward model.
\newblock \emph{arXiv preprint arXiv:2305.18290}.

\bibitem[{Ramamurthy et~al.(2022)Ramamurthy, Ammanabrolu, Brantley, Hessel, Sifa, Bauckhage, Hajishirzi, and Choi}]{ramamurthy2022reinforcement}
Rajkumar Ramamurthy, Prithviraj Ammanabrolu, Kiant{\'e} Brantley, Jack Hessel, Rafet Sifa, Christian Bauckhage, Hannaneh Hajishirzi, and Yejin Choi. 2022.
\newblock Is reinforcement learning (not) for natural language processing?: Benchmarks, baselines, and building blocks for natural language policy optimization.
\newblock \emph{arXiv preprint arXiv:2210.01241}.

\bibitem[{Roziere et~al.(2023)Roziere, Gehring, Gloeckle, Sootla, Gat, Tan, Adi, Liu, Remez, Rapin et~al.}]{roziere2023code}
Baptiste Roziere, Jonas Gehring, Fabian Gloeckle, Sten Sootla, Itai Gat, Xiaoqing~Ellen Tan, Yossi Adi, Jingyu Liu, Tal Remez, J{\'e}r{\'e}my Rapin, et~al. 2023.
\newblock Code llama: Open foundation models for code.
\newblock \emph{arXiv preprint arXiv:2308.12950}.

\bibitem[{Schulman et~al.(2017)Schulman, Wolski, Dhariwal, Radford, and Klimov}]{schulman2017proximal}
John Schulman, Filip Wolski, Prafulla Dhariwal, Alec Radford, and Oleg Klimov. 2017.
\newblock Proximal policy optimization algorithms.
\newblock \emph{arXiv preprint arXiv:1707.06347}.

\bibitem[{Stiennon et~al.(2020)Stiennon, Ouyang, Wu, Ziegler, Lowe, Voss, Radford, Amodei, and Christiano}]{stiennon2020learning}
Nisan Stiennon, Long Ouyang, Jeffrey Wu, Daniel Ziegler, Ryan Lowe, Chelsea Voss, Alec Radford, Dario Amodei, and Paul~F Christiano. 2020.
\newblock Learning to summarize with human feedback.
\newblock \emph{Advances in Neural Information Processing Systems}, 33:3008--3021.

\bibitem[{Yuan et~al.(2023)Yuan, Yuan, Tan, Wang, Huang, and Huang}]{yuan2023rrhf}
Hongyi Yuan, Zheng Yuan, Chuanqi Tan, Wei Wang, Songfang Huang, and Fei Huang. 2023.
\newblock Rrhf: Rank responses to align language models with human feedback.
\newblock In \emph{Thirty-seventh Conference on Neural Information Processing Systems}.

\bibitem[{Ziegler et~al.(2019)Ziegler, Stiennon, Wu, Brown, Radford, Amodei, Christiano, and Irving}]{ziegler2019fine}
Daniel~M Ziegler, Nisan Stiennon, Jeffrey Wu, Tom~B Brown, Alec Radford, Dario Amodei, Paul Christiano, and Geoffrey Irving. 2019.
\newblock Fine-tuning language models from human preferences.
\newblock \emph{arXiv preprint arXiv:1909.08593}.

\end{thebibliography}

\appendix

\section{Additional Hyper-parameter Experiments}
\begin{figure*}[tb]
    \centering
    \begin{subfigure}[b]{0.22\textwidth}
        \centering
        \includegraphics[width=\textwidth]{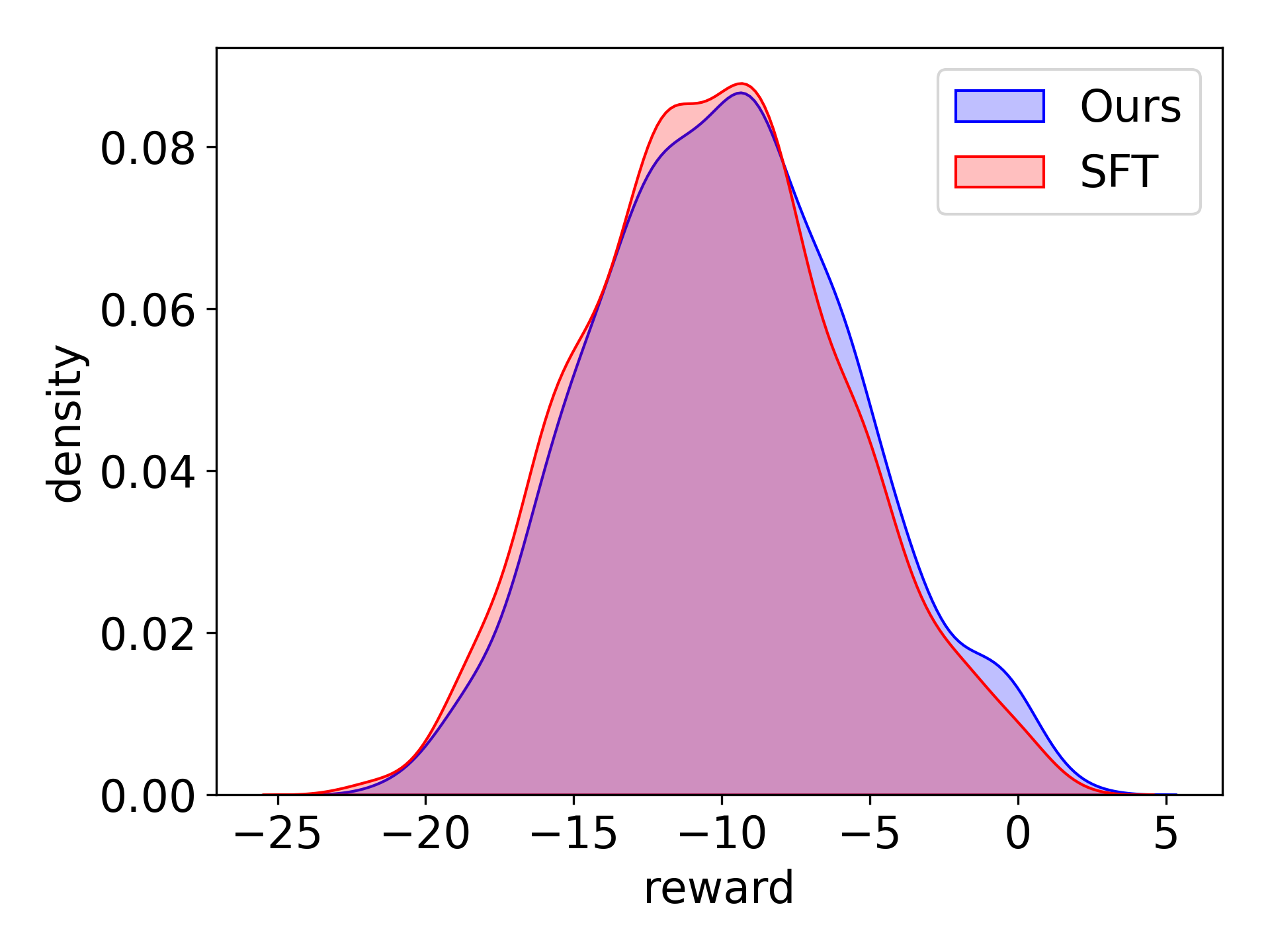}
        \caption{}
        \label{fig:temp1}
    \end{subfigure}
    \hfill
    \begin{subfigure}[b]{0.22\textwidth}
        \centering
        \includegraphics[width=\textwidth]{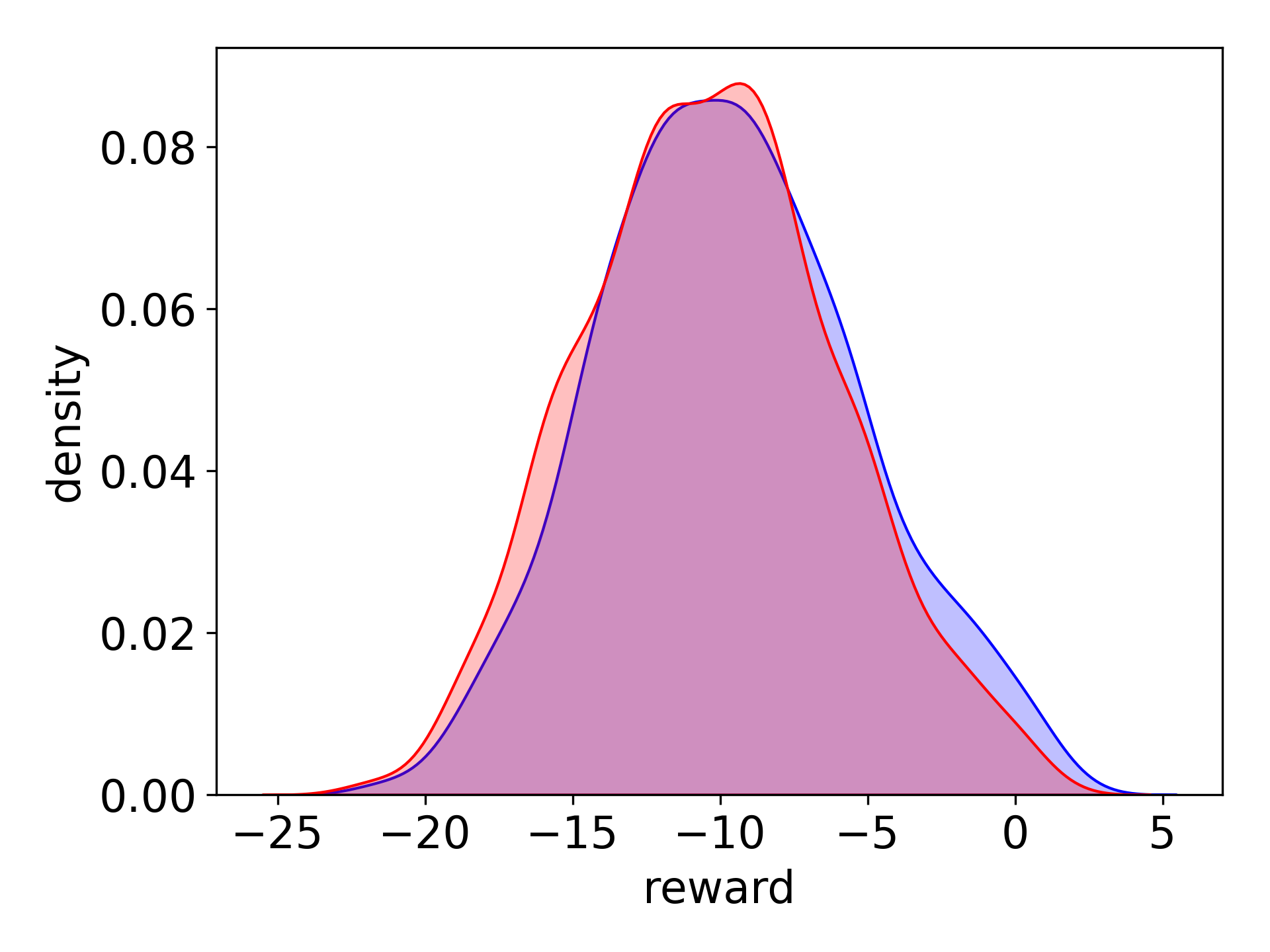}
        \caption{}
        \label{fig:temp2}
    \end{subfigure}
    \hfill
    \begin{subfigure}[b]{0.22\textwidth}
        \centering
        \includegraphics[width=\textwidth]{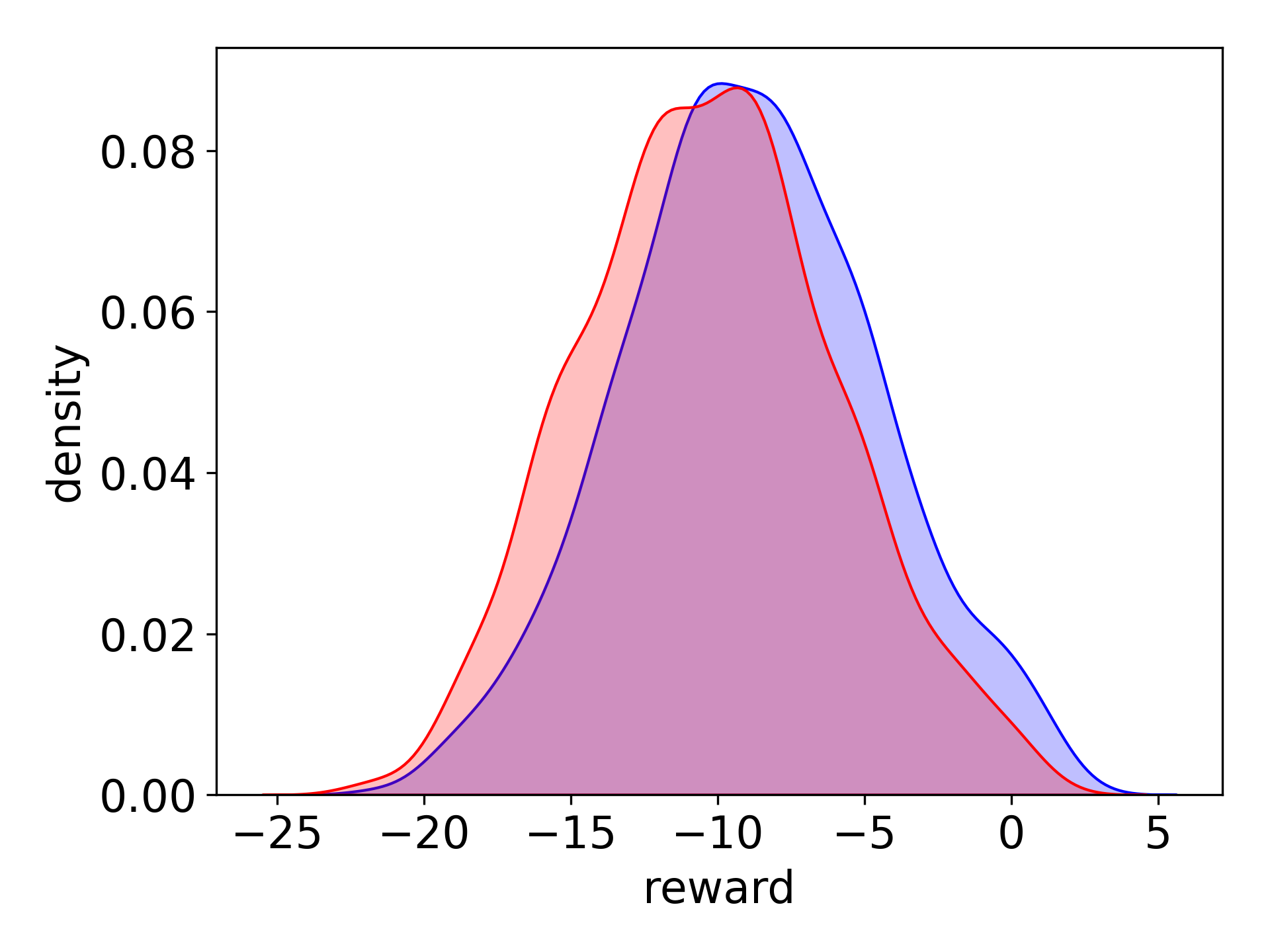}
        \caption{}
        \label{fig:temp3}
    \end{subfigure}
    \hfill
    \begin{subfigure}[b]{0.22\textwidth}
        \centering
        \includegraphics[width=\textwidth]{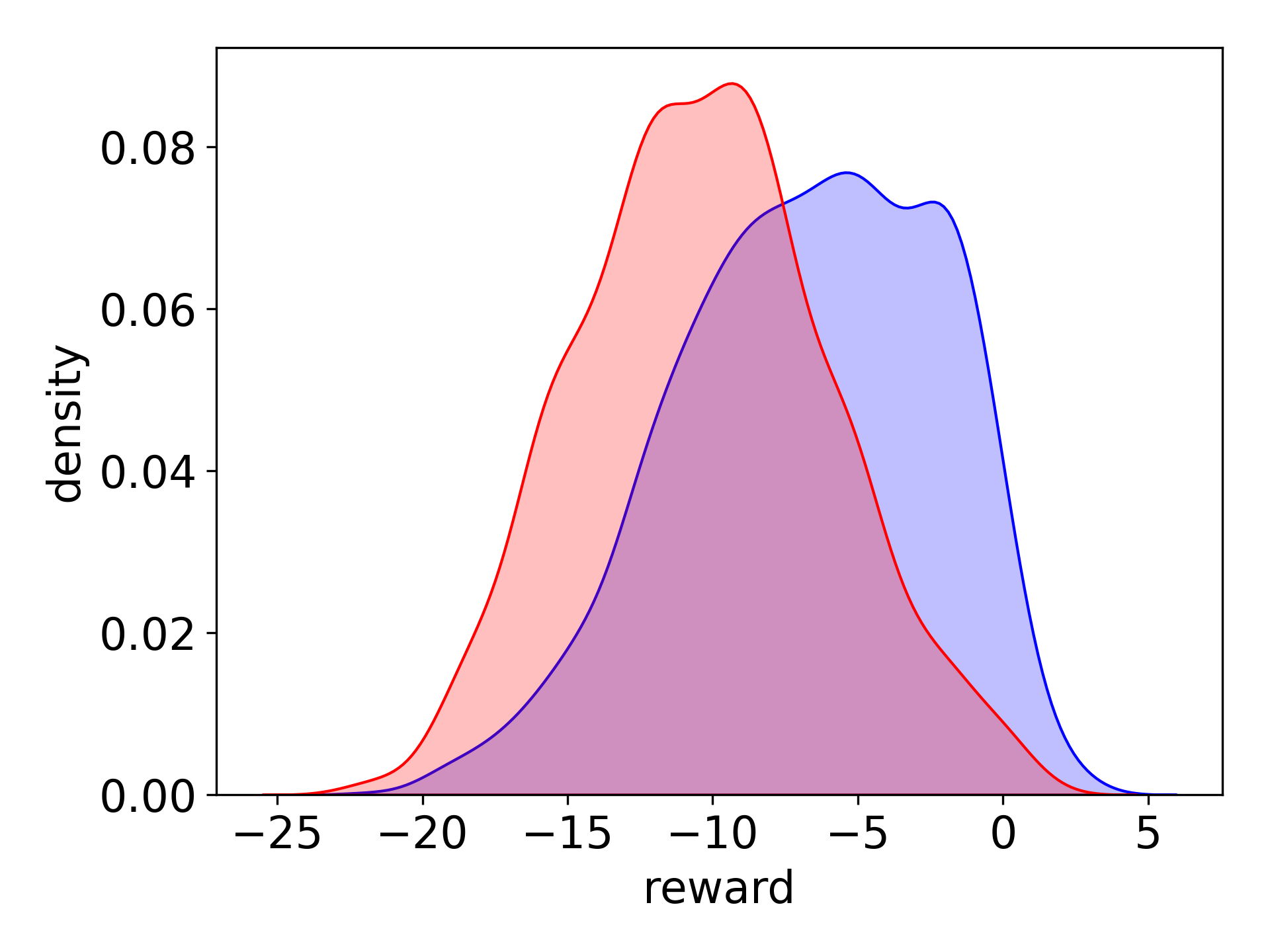}
        \caption{}
        \label{fig:temp4}
    \end{subfigure}

    \begin{subfigure}[b]{0.22\textwidth}
        \centering
        \includegraphics[width=\textwidth]{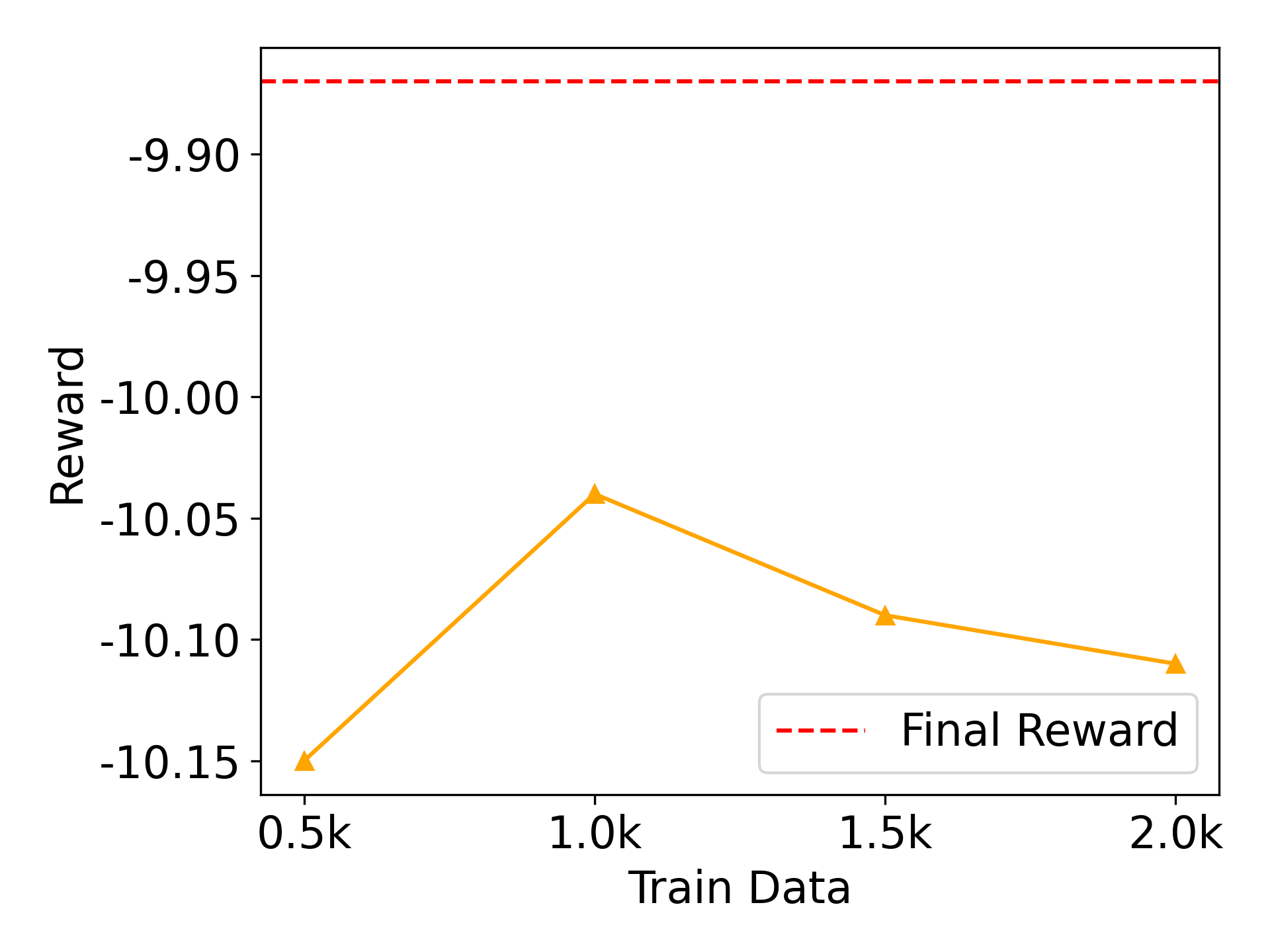}
        \caption{}
        \label{fig:temp5}
    \end{subfigure}
    \hfill
    \begin{subfigure}[b]{0.22\textwidth}
        \centering
        \includegraphics[width=\textwidth]{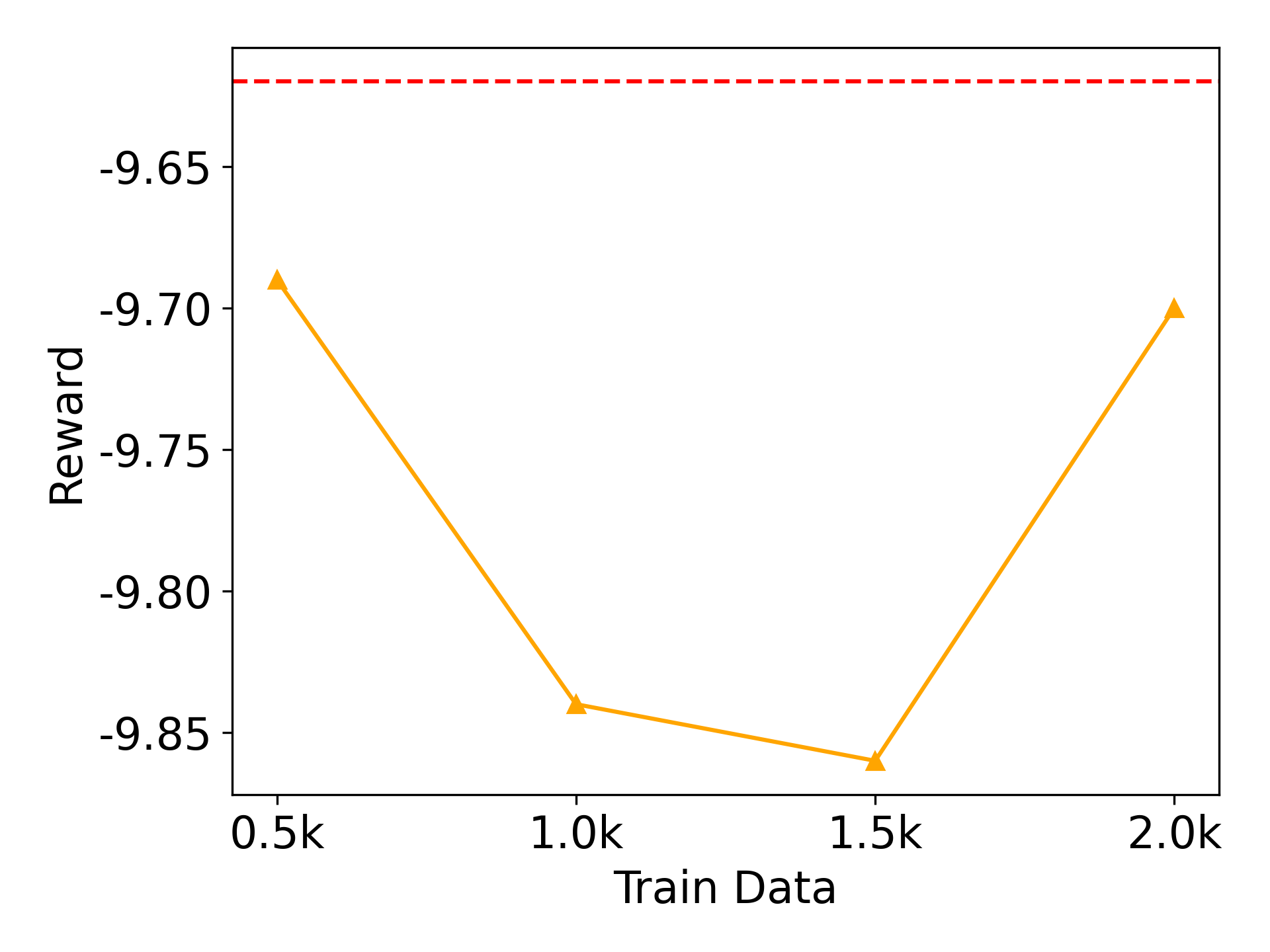}
        \caption{}
        \label{fig:temp6}
    \end{subfigure}
    \hfill
    \begin{subfigure}[b]{0.22\textwidth}
        \centering
        \includegraphics[width=\textwidth]{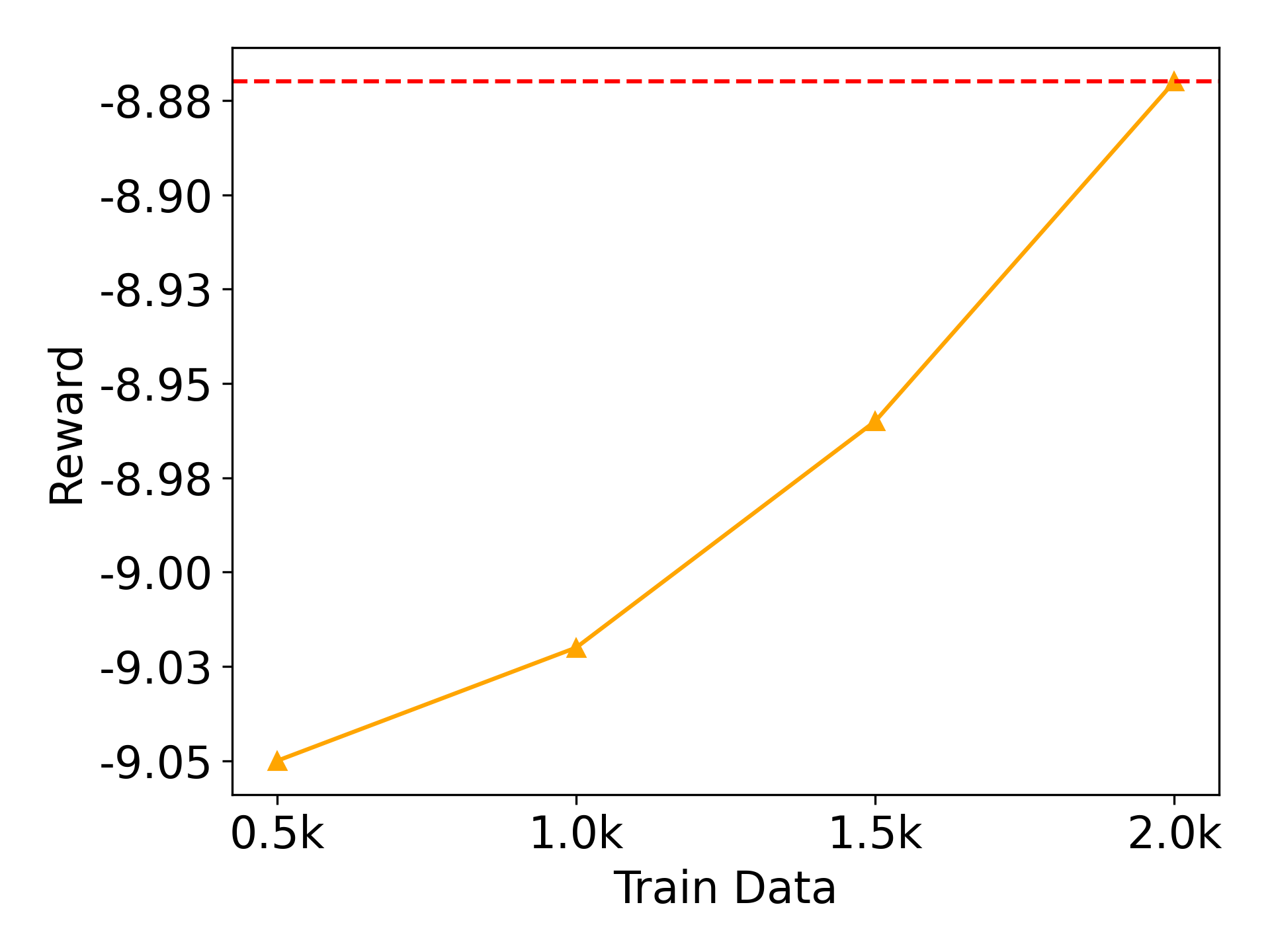}
        \caption{}
        \label{fig:temp7}
    \end{subfigure}
    \hfill
    \begin{subfigure}[b]{0.22\textwidth}
        \centering
        \includegraphics[width=\textwidth]{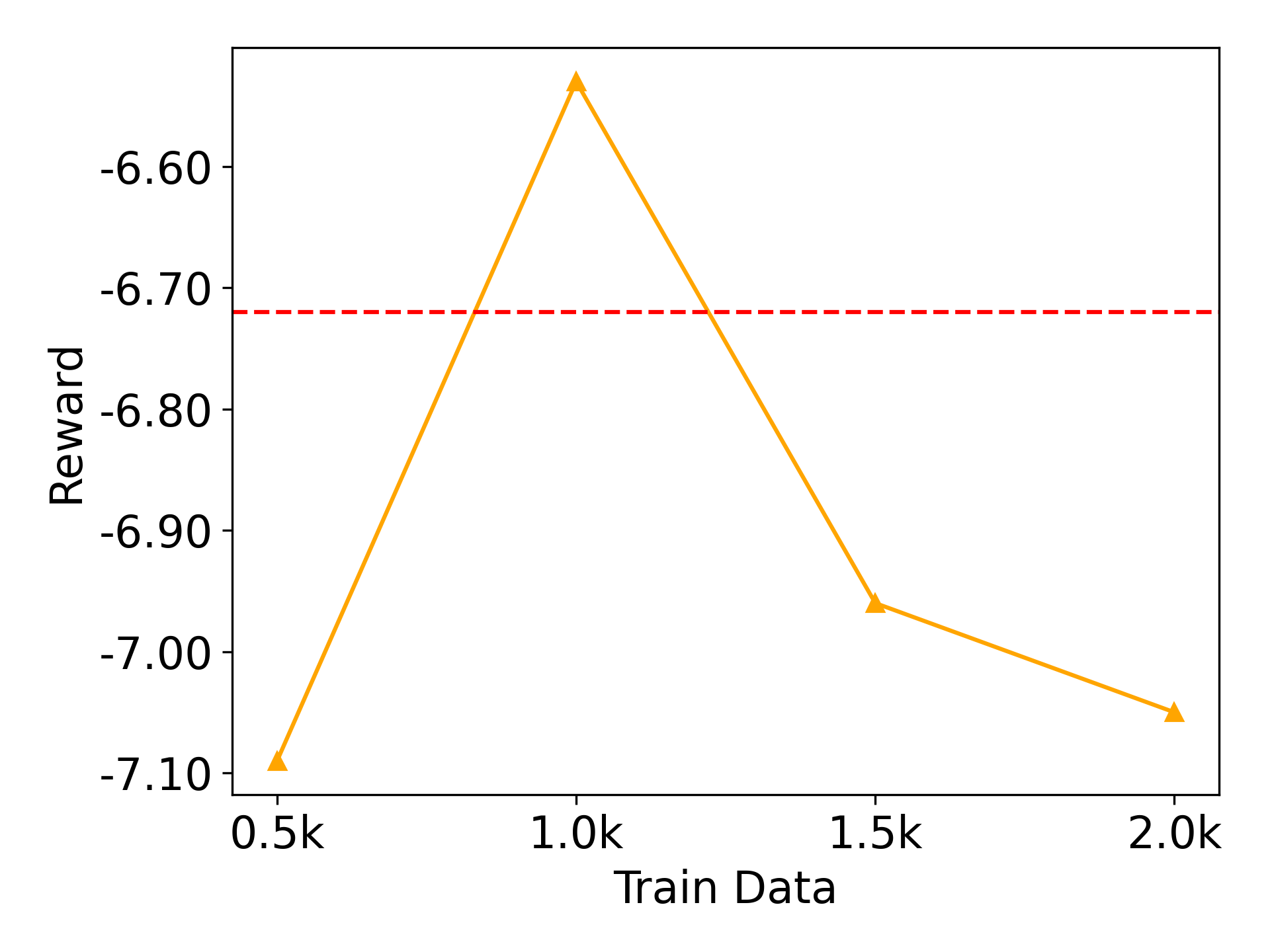}
        \caption{}
        \label{fig:temp8}
    \end{subfigure}
    
    \caption{(a)-(d) The reward distribution given by the reward model on the test set for the SFT model and models corresponding to different temperatures after completing one round on the training set. (a) temperature$=0.25$, (b) temperature$=0.5$, (c) temperature$=0.75$, (d) temperature$=1.0$. (e)-(h) The average score on the test set for models with different temperatures on the first 2k train data. (e) temperature$=0.25$, (f) temperature$=0.5$, (g) temperature$=0.75$, (h) temperature$=1.0$}
    \label{fig:1temp}
\end{figure*}

\begin{figure*}[htb]
    \centering
    \begin{subfigure}[b]{0.22\textwidth}
        \centering
        \includegraphics[width=\textwidth]{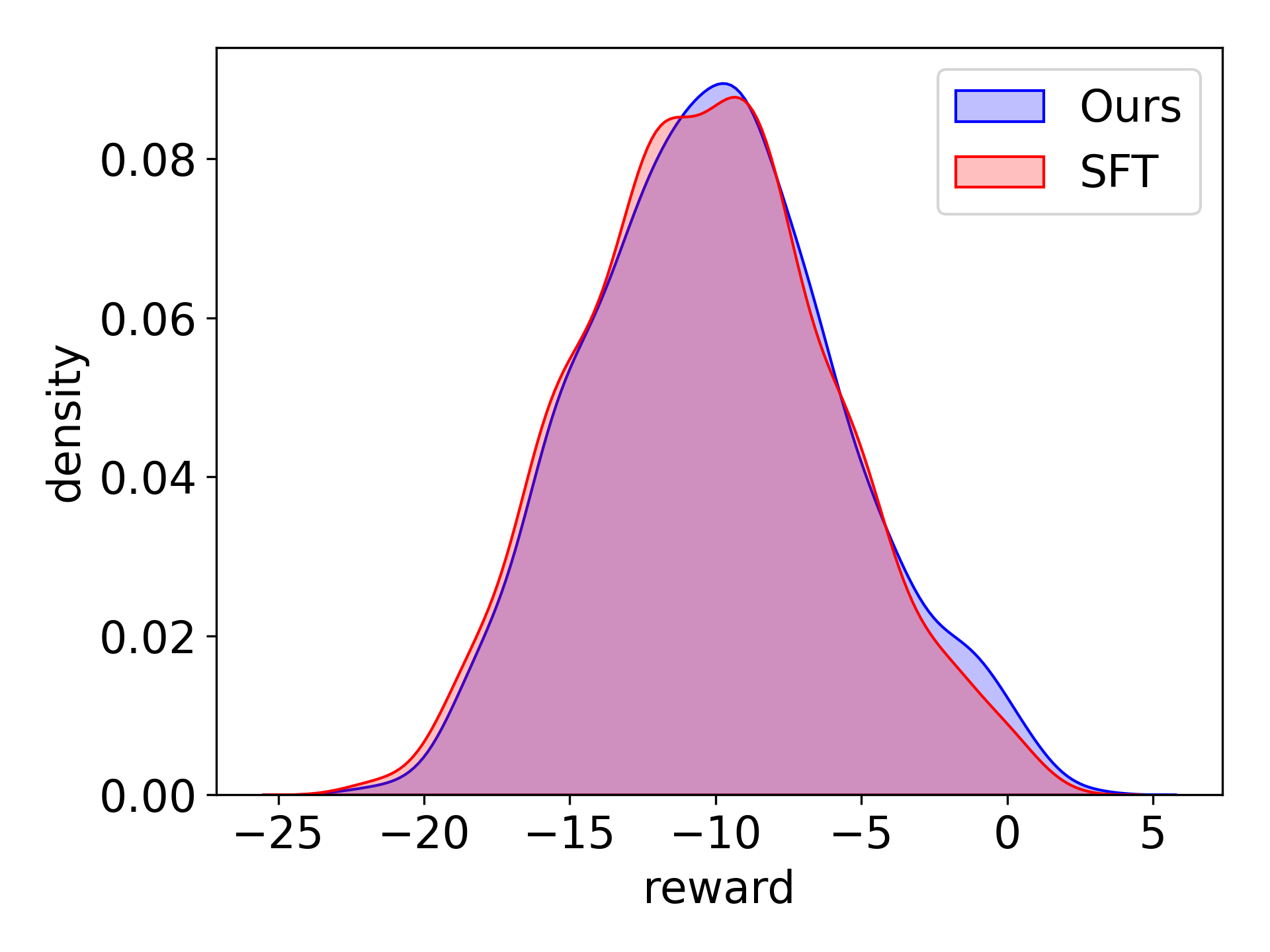}
        \caption{}
        \label{fig:pt1}
    \end{subfigure}
    \hfill
    \begin{subfigure}[b]{0.22\textwidth}
        \centering
        \includegraphics[width=\textwidth]{figures/t1_p01_fenbu_wl.png}
        \caption{}
        \label{fig:pt2}
    \end{subfigure}
    \hfill
    \begin{subfigure}[b]{0.22\textwidth}
        \centering
        \includegraphics[width=\textwidth]{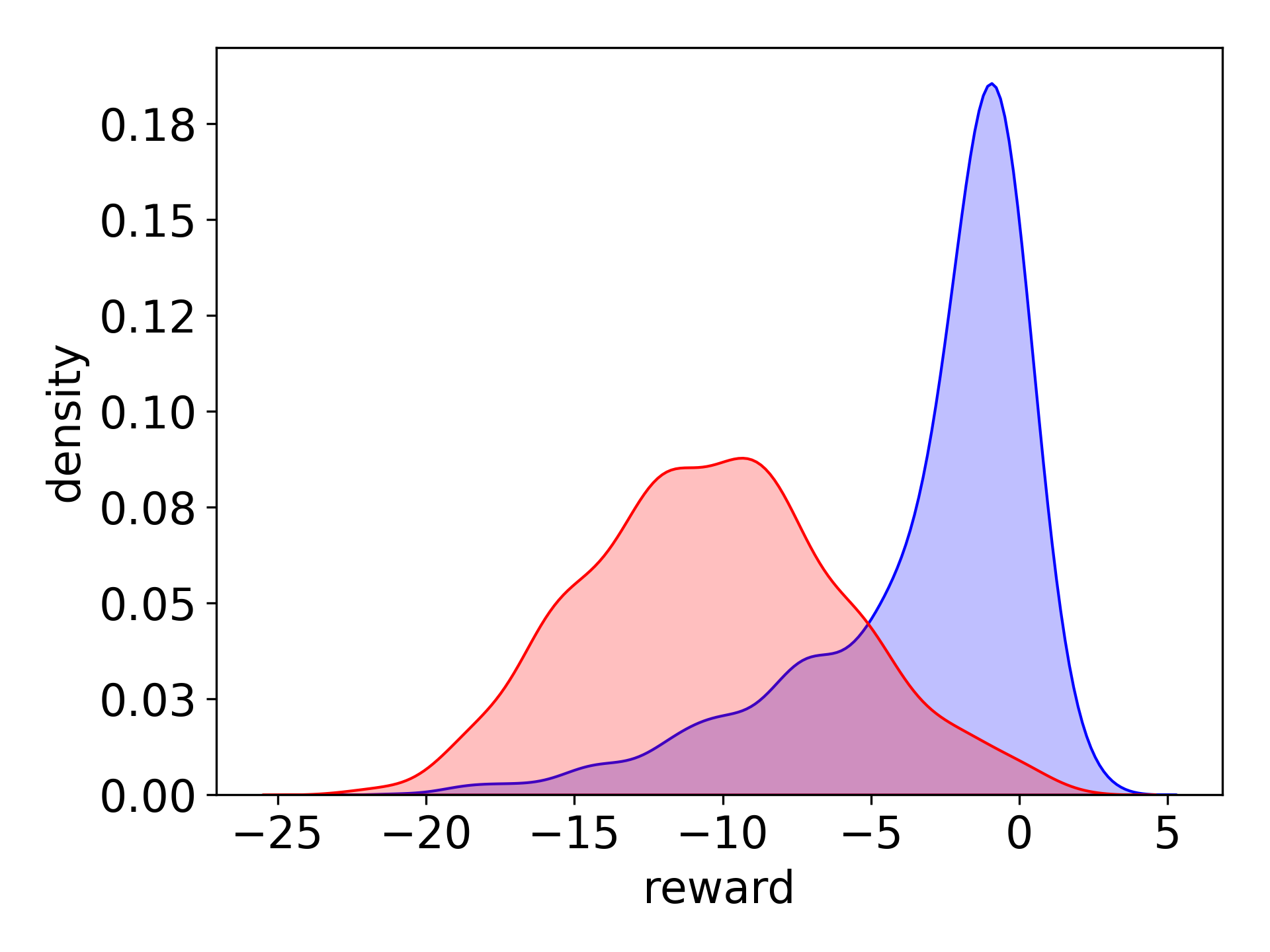}
        \caption{}
        \label{fig:pt3}
    \end{subfigure}
    \hfill
    \begin{subfigure}[b]{0.22\textwidth}
        \centering
        \includegraphics[width=\textwidth]{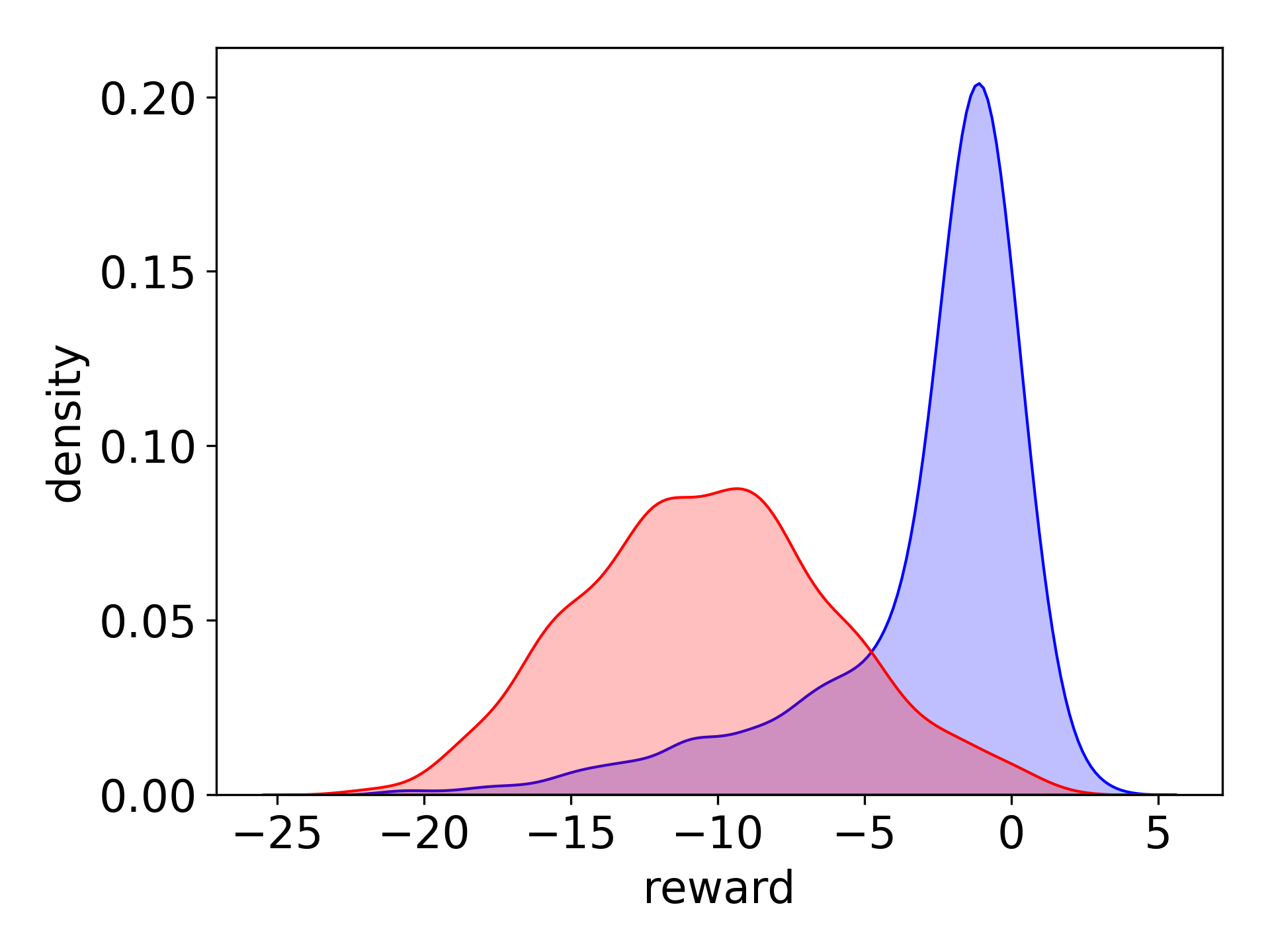}
        \caption{}
        \label{fig:pt4}
    \end{subfigure}

    \begin{subfigure}[b]{0.22\textwidth}
        \centering
        \includegraphics[width=\textwidth]{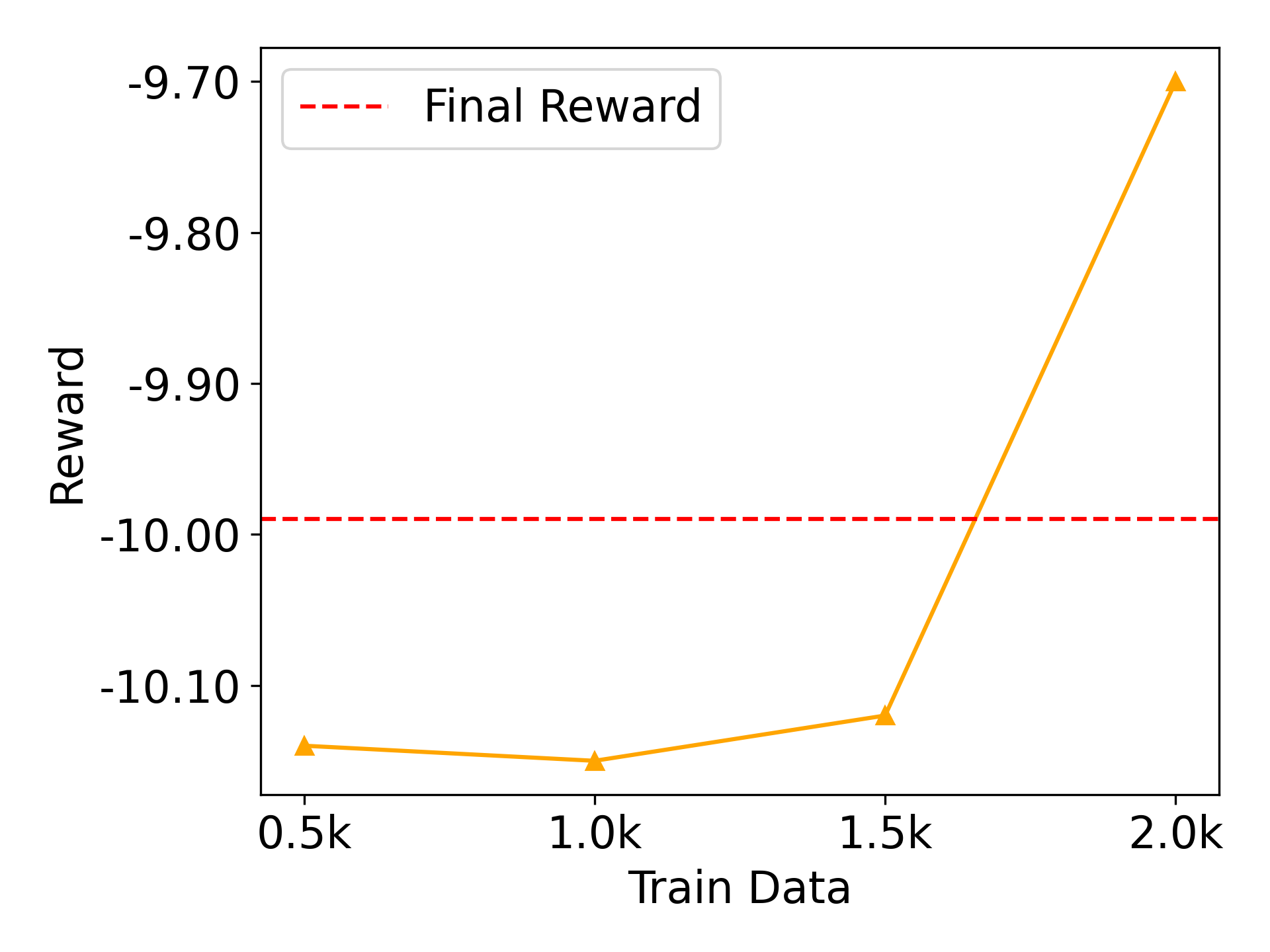}
        \caption{}
        \label{fig:pt5}
    \end{subfigure}
    \hfill
    \begin{subfigure}[b]{0.22\textwidth}
        \centering
        \includegraphics[width=\textwidth]{figures/t1_p01.png}
        \caption{}
        \label{fig:pt6}
    \end{subfigure}
    \hfill
    \begin{subfigure}[b]{0.22\textwidth}
        \centering
        \includegraphics[width=\textwidth]{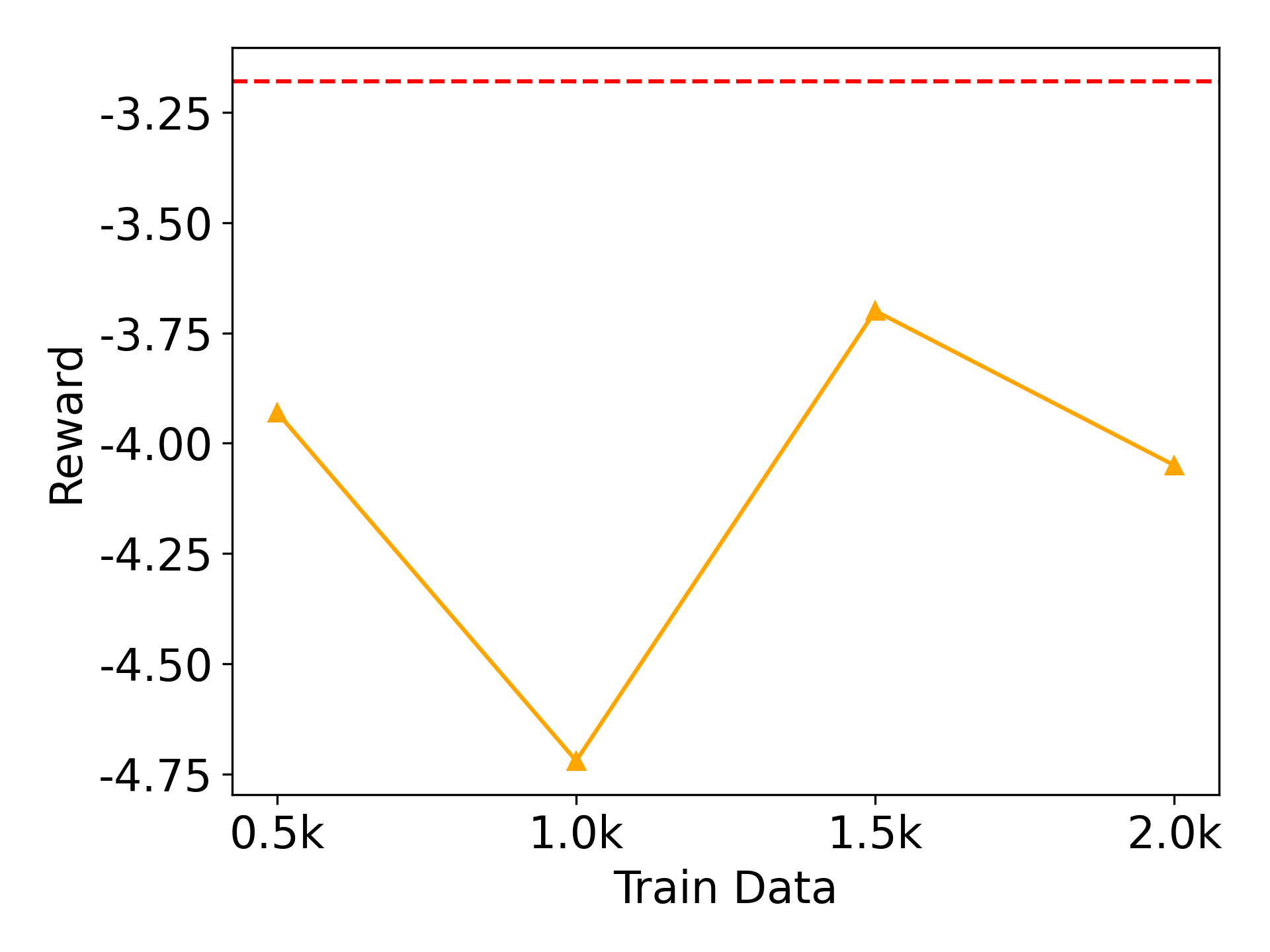}
        \caption{}
        \label{fig:pt7}
    \end{subfigure}
    \hfill
    \begin{subfigure}[b]{0.22\textwidth}
        \centering
        \includegraphics[width=\textwidth]{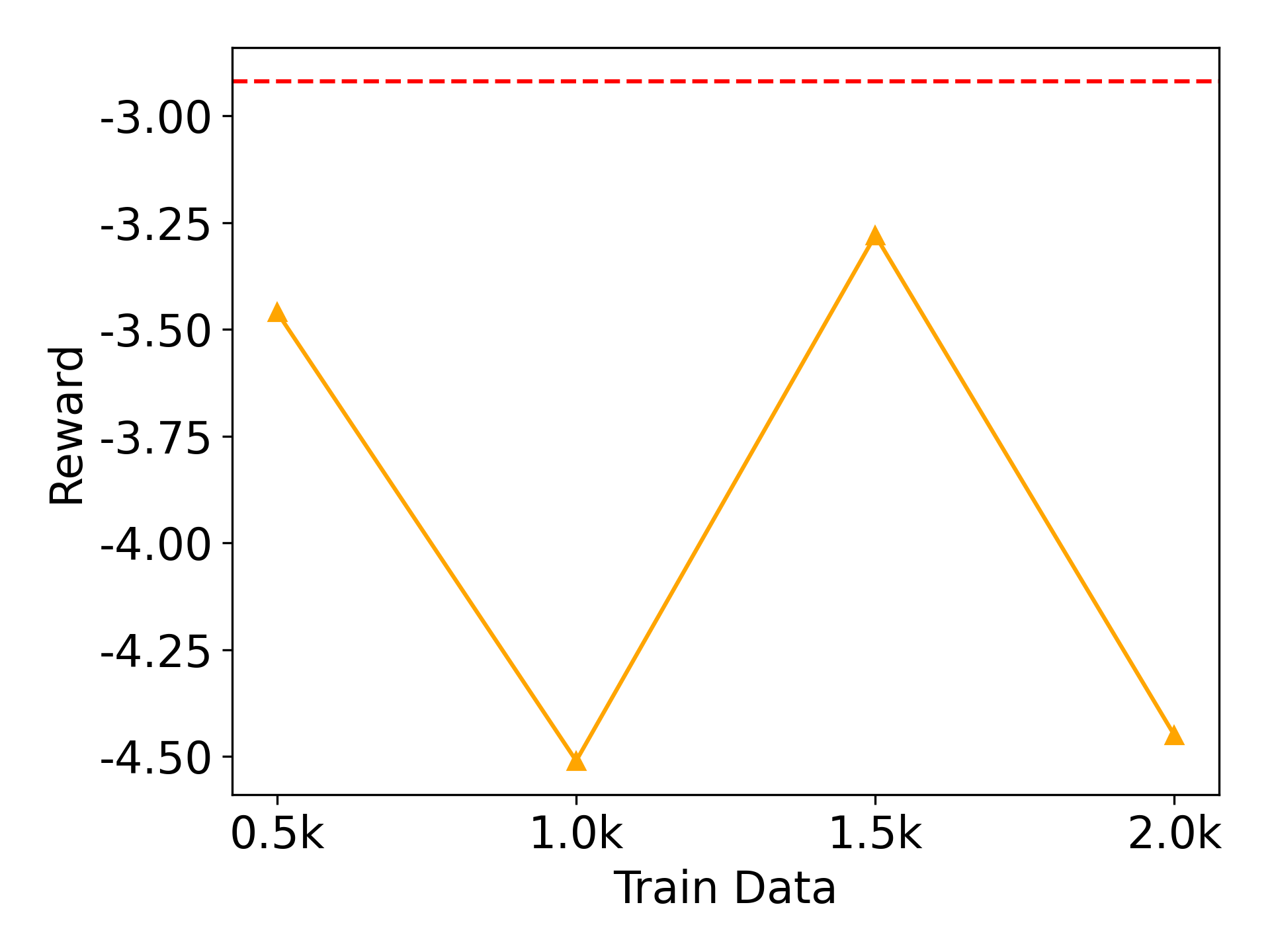}
        \caption{}
        \label{fig:pt8}
    \end{subfigure}
    
    \caption{(a)-(d) The reward distribution given by the reward model on the test set for the SFT model and models corresponding to different $p_t$ after completing one round on the training set. (a) $p_t=0.1$, (b) $p_t=0.01$, (c) $p_t=0.001$, (d) $p_t=0.0001$. (e)-(h) The average score on the test set for models with different $p_t$ on the first 2k train data. (e) $p_t=0.1$, (f) $p_t=0.01$, (g) $p_t=0.001$, (h) $p_t=0.0001$}
    \label{fig:1pt}
\end{figure*}

Figure \ref{fig:1temp} show that only with high temperature we can obtain a significant right-shift of reward distribution on the test set. However, for values of $p_t$, we can early achieve such improvement in Figure \ref{fig:1pt} when $p_t$ is reduced to 0.01.
In Figure \ref{fig:1temp} and Figure \ref{fig:1pt}, the red line in each subfigure indicates the average scores after one complete round of training. These figures show that the scores in early training steps are close to the final scores after one round of training. Especially for Figures \ref{fig:temp8},\ref{fig:pt5}, and \ref{fig:pt6}, we can achieve the final performance in early training steps no more than 2000.

\begin{table}[htb]
\centering
\caption{The Training Settings of Proxy-RLHF.}
\resizebox{\columnwidth}{!}{
\begin{tabular}{lc}
\toprule
Model    & 2-layer MLP     \\
Hidden Size    & 2048   \\
Learning Rate & 3e-4  \\
Computational Resource  & 2*NVIDIA A100 40G  \\
\bottomrule
\end{tabular}
}
\label{table:settings}
\end{table}

\section{Training Settings}
We show our experimental setting in Table \ref{table:settings}

\section{Irregular Response when $p_t$ is Too Small}

We list some irregular outputs including repeated tokens, irregular long words, etc. from experiment with $p_t=0.0001$ in Table \ref{table:prompt}.

\begin{table*}[hb]
\centering
\caption{Repeated tokens, irregular long words from experiment with $p_t=0.0001$.}
\begin{tabular}{|l|p{13cm}|}
\toprule
prompt             & USER: Recently heard about410
Joe Biden banning Huawei and ZTE from op-411
erating in the U.S. What are your thoughts412
on that?  \\
\midrule
response(Reward:-1.93)    & My thoughts on Joe Biden’s move of banning Huawei and ZTE from operating in the United States is a bittersweet. I applaud the effort as cyber security has to be prioritised. Nevertheless, U.S technology corporation releasing products internationally must also adhere to local cyber security measures imposed, regardless if American products can still penetrate deeply and widely likewise elsewhere due to patrons waking from sheepl. Hope both sides could still find ways on achieving cyber balancing by respecting multinational cyber norms which could safeguards critical national \& global digital security in long hauls. This is not war but responsible policing when every responsible Nation's effort on achieving Digital Peace should not rely  solely and entirely solely from the U.S cyber policeman's benevolent blessing in every single jurisdictional disputed Digital Battle Zones with very diverge Digital Cultures with very distinct Digital Norms which could not always synchronies to always adherent Universal Digital Norm. This could potentially put the entire Global Digital Infrastructural Security in perils if not properly tackles in a more more proactve cooperation between responsible Digital Partner in the Global Digital Playing Fileds in a more organise partnerships in the Global Digital Security Architecture in a more systemated coop in the Global Digital Security Architecture in a truly more more systemataive partnership in the Global Digital Security Network in Digital Age for Digital Peace and Digital Security in Digital Future for Digital Security in Digital Future for Digital Security in Digital Peace in Digital World for Digital Security in Digital Security in Digital Security in Digital Security in Digital Security in Digital Security in Digital Security in Digital Security in Digital Security in Digital Security in Digital Security in Digital Security in Digital Future in Digital Security in Digital Security in Digital Security in Digital Security in Digital Security in Digital Security in Digital Security in Digital Security in Digital Security in Digital Security in Digital Security in Digital Security in Digital Security in Digital Security in Digital Security in Digital Security in Digital Security in Digital Security in Digital Security in Digital Security in    \\

\bottomrule
\end{tabular}
\label{table:prompt}
\end{table*}

\end{document}